# Automatic Breast Ultrasound Image Segmentation: A Survey

Min Xian[1], Yingtao Zhang[3], H.D. Cheng[2,3], Fei Xu[2], Boyu Zhang[2], Jianrui Ding[3]

[1]*Department of Computer Science, University of Idaho, Idaho Falls, ID 83402, USA*

[2]*Department of Computer Science, Utah State University, Logan, UT 84322, USA*

[3]*School of Computer Science and Technology, Harbin Institute of Technology, China*

**Abstract**

Breast cancer is one of the leading causes of cancer death among women worldwide. In clinical routine, automatic breast ultrasound (BUS) image segmentation is very challenging and essential for cancer diagnosis and treatment planning. Many BUS segmentation approaches have been studied in the last two decades, and have been proved to be effective on private datasets. Currently, the advancement of BUS image segmentation seems to meet its bottleneck. The improvement of the performance is increasingly challenging, and only few new approaches were published in the last several years. It is the time to look at the field by reviewing previous approaches comprehensively and to investigate the future directions. In this paper, we study the basic ideas, theories, pros and cons of the approaches, group them into categories, and extensively review each category in depth by discussing the principles, application issues, and advantages/disadvantages.

## 1. Introduction

Breast cancer occurs in the highest frequency in women among all cancers, and is also one of the leading causes of cancer death worldwide [1, 2]. Scientists do not definitely know what causes breast cancer yet, and only know some risk factors that can increase the likelihood of developing breast cancer: getting older, genetics, radiation exposure, dense breast tissue, alcohol consumption, etc. The key of reducing the mortality is to find signs and symptoms of breast cancer at its early stage by clinic examination [3]. Breast ultrasound (BUS) imaging has become one of the most important and effective modality for the early detection of breast cancer because of its noninvasive, nonradioactive and cost-effective nature [4]; and it is most suitable for large-scale breast cancer screening and diagnosis in low-resource countries and regions.

Computer-Aided Diagnosis (CAD) systems based on B-mode breast ultrasound have been developed to overcome the considerable inter- and intra-variabilities of the breast cancer diagnosis, and have been clinically tested their ability to improve the performance of the breast cancer diagnosis. BUS segmentation, extracting tumor region of a BUS image, is a crucial step for a BUS CAD system. Base on the segmentation results, quantitative features will be calculated to describe tumor shape, size, echo pattern, etc., and be input into a classifier to determine the category of the tumors. Therefore, the precision of BUS segmentation directly affects the performance of the quantitative analysis and diagnosis of tumors. Segmentation is a common and crucial task in medical image analysis, and many medical image segmentation tasks share essentially similar segmentation approaches; however, comparing with medical image segmentation for other imaging modalities, e.g., computed tomography (CT), magnetic resonance imaging (MRI), mammography, BUS image segmentation is very challenging because (1) ultrasound image has very low quality due to the speckle noise, low contrast, low single noise ratio (SNR) and artifacts; (2) large variations of breast structures exists among patients, which make it difficult to apply the knowledge of anatomical structures; and (3) strong priors based on tumor shape, size and echo strength are important for organ segmentation [174, 180] in other imaging modalities; but these features vary largely across patients and are difficult to be applied to BUS image segmentation.

Automatic BUS segmentation has been studied extensively in the last two decades. We can classify existing approaches into semi-automated and fully automated groups according to the degree of human intervention involved in segmentation process. In most semi-automated methods, radiologist needs to specify a region of interest (ROI) including the lesion, a seed in the lesion, or an initial boundary. Fully automated segmentation approaches need no user intervention, and usually model the knowledge of breast ultrasound and oncology as the prior constraints. Many segmentation techniques were employed in both semi-automatic and fully automatic approaches.

[175] is the first survey paper for BUS image segmentation published recently; however, this paper is much more comprehensive in terms of major issue discussions, future direction prediction, theory fundamentals, application theme and the number of references. In this paper, we classify breast cancer segmentation approaches into six main categories: (1) graph-based approaches [7, 9-13, 22-25, 27, 28, 32-34], (2) deformable models [42, 50, 52, 54-59, 61-65, 67, 69, 71, 73, 75, 76, 78-82, 85, 124, 139, 148, 166], (3) learning-based approaches [7, 9,10, 87, 89, 91, 94, 95, 98,100, 103-107, 120], (4) thresholding [22, 109-115], (5) region growing [54, 55, 113, 117, 118], and (6) watershed [109, 122, 123, 126-128]. As shown in Fig. 1, the first three categories dominate BUS image segmentation approaches; and the last three categories are the classical image

processing approaches. The category of the others [136, 138, 139, 140, 142-146] is composed of three small sub-categories, each contains only few literatures. Due to the challenging nature of the task, just using single image processing technique cannot achieve desirable results; and most successful approaches employ hybrid techniques and model biological priors.

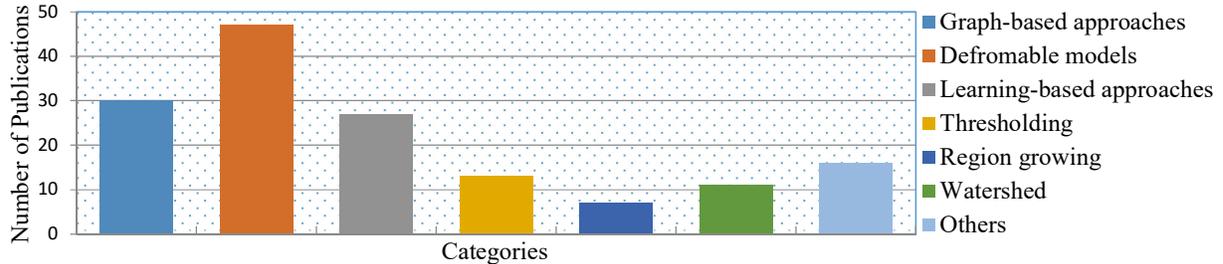

Fig. 1. Distribution of automatic BUS image segmentation approaches; data come from google scholar until May 2017.

The rest of the paper is organized as follows: in section 2, the fundamental issues in BUS segmentation are discussed, e.g., denoising, interaction, biological priors modeling, validation, and the possible problem-solving strategies; in sections 3 - 6, we review automatic BUS image segmentation methods by presenting the principle of each category, discussing their advantages and disadvantages, and summarizing the most valuable strategies. In section 7, we discuss the approaches of three sub-categories briefly. Section 8 gives the conclusion and the future directions.

**2. Fundamental Issues of BUS Image Segmentation**

BUS segmentation approaches have been studied in the last two decades extensively, and many of them achieved good performances utilizing their own datasets. In this section, we discuss the fundamental issues in BUS segmentation, and summarize the successful strategies employed in state-of-the-art approaches.

**2.1 Denoising and Preserving Edge**

In ultrasound imaging, speckle noise is inherent to coherent illumination and Rayleigh scattering caused by tissue microstructures [147]; and it is a major difficulty in BUS image segmentation because the speckle artifacts are tissue-dependent and cannot be effectively modeled [167]. Many de-speckle approaches have been applied, e.g., mean filter, Gaussian low-pass filter, speckle reducing anisotropic diffusion (SRAD) [148], non-linear coherence diffusion (NCD) [149], sticks filter [60], bilateral filter [150, 151], fractional subpixel diffusion [157], nonlocal means-based speckle filter [167], etc. Mean filter and Gaussian low-pass filter are simple, fast, and easy to implement. They are widely used in early BUS segmentation approaches [22, 52, 59, 61-63, 109, 112, 115, 123, 133, 138]; SRAD, NCD, sticks filter, and fractional subpixel diffusion are specially de-signed to deal with speckle noise, and because of their excellent property in edge preservation, these approaches are employed in BUS segmentation popularly [34, 58, 65, 89, 95, 103, 105, 106, 113, 114, 124, 125, 158, 165, 166]. The mean filter and Gaussian low-pass filter have the side effect of blurring edges and are only suitable for the approaches insensitive to image edges or gradient. For the edge-based approaches, e.g., edge-based deformable models and watershed, denoising with edge preservation approaches are preferred to avoid the leakage of final segmentation. For more details about modern image denoising or filtering approaches, please refer [168].

**2.2 Human Intervention**

Many semi-automatic approaches exist in literature; and user interactions like setting seeds, drawing initial boundary or ROI are required in these approaches. Radiologists' interactions could be useful in segmenting extremely difficult BUS cases which have very poor image quality; however, these interactions make the approaches operator-dependent and the results unreproducible; furthermore, it is also impossible to apply the semi-automatic approaches to a large-scale BUS image dataset, because of the great cost of human labor and time. The intensity and sensitivity of interaction are two important criteria for evaluating interactive segmentation approaches [152, 153]; because user interaction has large degree of arbitrariness and different interaction may lead to quite different results. However, no work has been done in BUS image segmentation to evaluate the sensitivity and intensity of the approach's interaction yet.

Fully automatic BUS image segmentation has many advantages in comparison with interactive segmentation, such as fully operator-independent, reproducible, and suitable for large scale tasks; therefore, we believe that fully automatic segmentation is the trend in the future BUS CAD systems. Many fully automatic approaches [22, 25, 64, 98, 100, 106, 109, 115, 154] have been proposed in the last few years. *In fully automatic BUS segmentation, the key step to make an approach completely automatic is the tumor detection, which outputs ROI, rough boundary, seeds, or candidate regions to localize tumors, and initializes the subsequent segmentation steps*. The approaches [54, 55, 106, 113] constructing empirical formulas to model domain priors provided effective means for tumor detection, but the predefined reference point (RP) formulated in these approaches limited the robustness and flexibility. One direction to improve these approaches is to model more robust prior, e.g., adaptive RP [22]; the second direction is to improve the generality of the formulas by constructing them in a learning-based framework. Learning based fully automatic approaches [25, 64, 98, 100] are promising and increasing popularity recently. There are two directions to improve these approaches: (1)

incorporating both global and local features into the learning framework; and (2) learning deep representation of breast structure towards the better understanding of BUS image by using deep convolutional neural network.

**2.3 Modeling Prior Knowledge**

Many ordinary image segmentation frameworks have been applied to BUS image segmentation; however, just applying ordinary image segmentation approaches cannot achieve good performance; and successful BUS segmentation approaches should model domain-related priors appropriately. We summarize the major priors that have been modeled in BUS image segmentation as follows.

*Intensity distribution*. It is widely used in BUS image segmentation, and the approaches can be classified into following classes: (1) using empirical distribution to model the intensity distribution of tumor or normal tissues, e.g., Gaussian, Raleigh, exponential model, etc; and (2) defining intensity distribution implicitly by using histograms and classifiers. In graph-based approaches [7, 9 - 11], Gaussian distribution of tumor intensity was usually applied to define the likelihood energy (data term). Liu et al. [67] modeled the probability density difference between the intensity distribution of tumor/background region and estimated Rayleigh distribution to improve the performance of Geometric Deformable Model (GDM). In [22, 54, 55], no explicit distribution was predefined, and only histogram was applied to describe the distribution of tumor region and normal tissues. In [25, 27, 28], supervised learning approaches are introduced to train the classifiers to output the probability of each image region to be tumor or background.

*Texture and local region statistics*. Texture and other local region features have more descriptive power than intensity, and have been studied in many works [9, 10, 64, 98]. They can distinguish tumor regions from normal tissues with high accuracy. In [9, 10], the texture distributions were utilized to build the likelihood energy of the graph model. Madabhushi et al. [54, 55] trained the texture histogram of tumor regions, and incorporated it with the trained intensity distribution to determine the candidate tumor regions. Liu et al. [64, 98] extracted statistic texture from local regions ($16\times 16$ grid), and learned a SVM classifier to localize tumors accurately.

*Edge or gradient*. In edge-based deformable models, [42, 50, 52, 56 - 59, 61 - 63], image gradient is applied to constructing the external force or speed function of the evolving curve; as discussed in section 4.3, because of the speckle noise and week boundary problems, the performance of most approaches depend on both denoising and edge preservation techniques; [73, 75] defined the stop function of GDM for edge detection in the frequency domain rather than in the spatial domain, which made the GDM insensitive to image contrast and work well on weak boundaries. Xian et al. [22, 115] proposed an edge detector in the frequency domain and incorporated it into a graph-based framework, which made the pairwise energy less sensitive to image contrast and brightness.

*Layer structure*. The breast is composed of different layers of tissues, e.g., skin, premammary, mammary, retromammary, and muscle layers [169]. Due to the difference of their physical properties, different layers have different appearances in BUS images. The location and depth of these layers may have big variations and they are difficult to detect; however, some works [22, 115, 154] had utilized the layer information for segmenting tumors.

*Topological properties*. Human vision system is very sensitive to the topological properties of objects, and some works have been investigated [115, 152, 153] for both natural and medical image processing tasks. In [152, 153], the Neutro-Connectedness (NC) is proposed to compute both the connectedness structure and map, which has solved the problems of high interaction intensity and interaction-dependence in interactive image segmentation successfully.

*Smoothness*. In graph-based models, it corresponds the smoothness term (pairwise energy), and the minimization of the energy makes the models produce a smooth boundary; however, it is important to notice that the smoothness term also makes the models have the tendency to shrink and generate much shorter boundary than the real boundary ("Shrink" problem [172]).

**2.4 Validation: ground truth, metrics, and BUS image datasets**

Most BUS image segmentation approaches discussed in this paper have been evaluated quantitatively, and two major approaches have been utilized. The first approach is to evaluate segmentation performance by using physical phantoms or simulation software, e.g., Field II [155, 156]. The advantage of this approach is that the ground truth is very accurate; but the physical phantoms cannot represent the real breast anatomy exactly, and the simulation software uses simple acquisition model and cannot represent the real BUS image acquisition process well. The second approach for validation is to compare the segmentation result with manually delineated boundary. The problem of this approach is that the manually delineated ground truth could be flawed because of human error; however, this problem could be solved by labeling the boundary multiple times by same person and/or multiple radiologists. Currently, evaluating BUS image segmentation approaches through the ground truth delineated by radiologists is widely accepted.

In Table 1, we list eight commonly used metrics; the first six are area metrics and the last two are boundary metrics. Notice that in some papers [22, 23, 115, 127, 137], the similarity index (SI) is defined as Jaccard index (JI); while in paper [61], SI is defined as the Dice's coefficient (DSC). FNR equals to (1 – TRP); therefore, it is not necessary to compute both.

Currently, there is no public BUS image benchmark with a reasonable number of clinical cases; the performance of most approaches are validated by only using the private datasets; and it is difficult to compare different approaches objectively. Therefore, there is a pressing need for establishing a BUS image benchmark. It will be valuable for comparing the algorithms by using a public dataset objectively, and for determining which approach achieves better performance and what segmentation strategies should pursue.

Table 1. Quantitative metrics for BUS image segmentation

| Metrics | Definition | Alias in References |
| --- | --- | --- |
| True positive ratio (TPR) | $\|R \cap G\|/\|G\|$ | Recall rate, overlap fraction, |
| False negative ratio (FNR) | 1 - TPR | None |
| False positive ratio (FPR) | $\|R \cup G - G\|/\|G\|$ | Error fraction |
| Jaccard index (JI) | $\|R \cap G\|/\|R \cup G\|$ | Similarity Index, coincidence percentage |
| Dice's coefficient (DSC) | $2\|R \cap G\|/(\|R\| + \|G\|)$ | Similarity Index, Dice similarity |
| Area error ratio (AER) | $\|(R \cup G) - (R \cap G)\|/\|G\|$ | Difference radio, normalize residual value |
| Hausdorff distance (HD) | $\max\{\max_{x \in R}\{d(x,G)\}, \max_{y \in G}\{d(y,R)\}\}$<br>$d(x,C) = \min_{y \in C}\{\|x-y\|\}, C = R \text{ or } G$ | None |
| Mean absolute distance (MAD) | $1/2(\sum_{x \in R} \frac{d(x,G)}{N_R} + \sum_{y \in G} \frac{d(y,R)}{N_G})$ | Mean error distance |

## 3. Graph-based approaches

Graph-based approaches gain increasing popularity in BUS image segmentation due to their advantages: (1) they provide a simple way to organize task-related priors and image information in a unified framework; (2) they are flexible and suitable for expressing soft constraints among random variables; and (3) the computation based on graphical manipulation is very efficient [5].

Let $G = (\mathcal{V}, \mathcal{E})$ be a graph comprising a set of nodes (vertices) $\mathcal{V} = \{v_1, v_2, \cdots, v_n\}$, and each of them corresponds to an image pixel/superpixel; and a set of links (edges) $\mathcal{E} = \{\langle v_i, v_j \rangle | v_i, v_j \in \mathcal{V}\}$, and each of them connects two adjacent nodes according to a predefined neighborhood system $\mathcal{N} = \{N_i | i = 1, \cdots, n\}$ where $N_i$ is the set of neighbors of node $v_i$. Each link $\langle v_i, v_j \rangle$ is associated with a nonnegative weight $w\langle v_i, v_j \rangle$. The weight is usually defined as the cost of separating the two connected nodes into different classes. For state-of-the-art superpixel generation approaches, refer [177-179].

### 3.1 MRF-MAP approaches

Markov random field (MRF) is an undirected graphical model and provides a convenient way to model context-dependent entities (pixels/superpixels). In MRF, a site set $S = \{i\}_{i=1}^n$ indexes the node set $\mathcal{V}$; each site $i$ is associated with a random variable $X_i$; $x = \{x_i\}_{i=1}^n$ is the configuration (implementation) of random variable set $X = \{X_i\}_{i=1}^n$, and $x_i$ takes value from a label set $\mathcal{L} = \{l_i\}_{i=1}^m$ where $m$ is the number of labels (classes). In BUS image segmentation, the label set is usually defined as $\mathcal{L} = \{l_1, l_2\}$ where $l_1$ denotes tumor and $l_2$ denotes non-tumor.

Let $p(x)$ be the joint probability (also called the prior distribution) denoted as $p(X = x)$. $X$ is said to be a MRF on $S$ with respect to a neighborhood system $\mathcal{N} = \{N_i\}_{i=1}^n$ if and only if it satisfies the positivity and Markovianity:

$$p(x) = p(x_i|x_{S-\{i\}})p(x_{S-\{i\}}) > 0$$
$$= p(x_i|x_{N_i})p(x_{S-\{i\}}) \quad (1)$$

where $x_{S-\{i\}}$ denotes a set of labels for sites $S - \{i\}$, and $x_{N_i}$ is the set of labels for site $i$'s neighbors ($N_i$).

Maximum a posteriori (MAP) is the most popular optimality criterion for MRF modeling and the optimal ($x^*$) is found by

$$x^* = \underset{x}{\mathrm{argmax}}\, p(x|d) = \underset{x}{\mathrm{argmax}}\, p(d|x)p(x) \quad (2)$$

where $d$ is the observation (image), $p(x|d)$ is the posterior distribution and $p(d|x)$ is the likelihood distribution. The Hammersley-Clifford theorem [6] established the equivalence between MRF and Gibbs random field (GRF), and the MAP is equivalently found by minimizing the posterior energy function

$$x^* = \underset{x}{\mathrm{argmin}}\, E(x|d) = \underset{x}{\mathrm{argmin}}\bigl(E(d|x) + E(x)\bigr) \quad (3)$$

where $E(x|d) = E(d|x) + E(x)$, and $E(d|x)$ is the likelihood energy and $E(x)$ is prior energy.

There are two major parts in the MRF-MAP modeling for BUS image segmentation: (1) defining the prior and likelihood energies and determining the corresponding initial parameters; and (2) designing optimization algorithm for finding the minimum of the posterior energy. Ashton and Parker [7] defined the MRF prior energy as the Potts model

$$E(x) = \sum_{i \in S} V(x_i) + \sum_{i \in S} \sum_{j \in N_i} V(x_i, x_j) \quad (4)$$

$$V(x_i, x_j) = \begin{cases} -\beta, & \text{if } x_i = x_j \\ \beta, & \text{otherwise} \end{cases} \quad (5)$$

where $\beta$ is a positive constant. The assumption that every site $s$ takes any label equally likely makes $V(x_i)$ a constant for all configurations; therefore, $E(x)$ is usually defined only on the pairwise term ($V(x_i, x_j)$). They also assume that the intensities of a low-pass filtered image follow the Gaussian distribution, and the likelihood energy is given by

$$E(d|x) = \sum_{i \in S} \left[ \ln(\delta_i^{x_i}) + \frac{(d_i - \mu_i^{x_i})^2}{2(\delta_i^{x_i})^2} \right] \quad (6)$$

where $d_i$ is the intensity of the pixel at site $i$, $x_i$ is the label of site $i$, and $\mu_i^{x_i}$ and $\delta_i^{x_i}$ are the local class mean and standard deviation, respectively. The parameters ($\mu_i^{x_i}$ and $\delta_i^{x_i}$) are estimated by using a modified adaptive k-mean method [8].

Boukerroui et al. [9] improved the method in [7] by modeling both intensity and texture distributions in the likelihood energy. Boukerroui et al. [10] modified the method in [9] by introducing a weighting function considering both global and local statistics. [12] improved the approach in [11] by introducing one-click user interaction to estimate Gaussian parameters automatically. [13] introduced the tissue stiffness information of ultrasound elastography to the method in [11] by modifying one-dimensional tumor and background Gaussian distributions as bivariate Gaussian distributions.

The energy function of MRF-MAP can be optimized by using Simulated Annealing (SA) [14] and Iterated Conditional Mode (ICM) [15] algorithms. Because ICM is much faster than SA, the ICM is preferred in most BUS image segmentation approaches [7, 9, 11 - 13]. ICM takes a local greedy strategy: it starts with an initial labeling from estimation or user interaction, then selects label for each site to minimize the energy function; and the steps repeat until convergence. ICM is quite sensitive to the initialization (parameters) because of its high possibility of converging to local minima, especially, for non-convex energies in high-dimensional space [35]. A detailed comparison between ICM and other MRF energy minimization techniques can be found in [35]. Therefore, one important issue of applying ICM is how to learn the parameters: Gibbsian parameter $\beta$ (Eq. (5)), number of classes ($m$), and parameters of the initial Gaussian distribution. In BUS image segmentation, the number of classes is usually set from 2 to 4; $\beta$ could be set adaptively [7] or set to a constant [11-13]; and the parameters of Gaussian distribution are usually initialized by using K-means algorithm [7, 9, 10].

### 3.2 Graph cuts

Graph cuts was proposed to solve a special case in MRF-MAP framework [16]: the global optimization of the binary labelling problem ($\mathcal{L} = \{l_1, l_2\}$). It was then improved to solve the general (color image segmentation) and multi-labeling problems [17, 18]. The main idea of graph cuts is to employ the theories and algorithms of the min-cut (*s-t* cut) and max-flow for binary segmentation. The graph $G^{gc}$ is defined as following.

$$G^{gc} = (\mathcal{V}^{gc}, \mathcal{E}^{gc}), \mathcal{V}^{gc} = \mathcal{V} \cup \{s, t\},$$
$$\mathcal{E}^{gc} = \underbrace{\mathcal{E}}_{n-lin} \cup \underbrace{\{\langle v, s \rangle | v \in \mathcal{V}\} \cup \{\langle v, t \rangle | v \in \mathcal{V}\}}_{t-links} \quad (7)$$

As shown in Eq. (7), graph cuts introduces two auxiliary nodes $s$ and $t$ representing the source and sink, respectively; the newly added edges between each node in $\mathcal{V}$ and the terminal nodes ($s$ and $t$) are called the *t-links*, and the original edges among neighboring nodes in $\mathcal{E}$ are called the *n-links*. A cut ($C^{gc}$) of $G^{gc}$ is defined by

$$C^{gc} = \{\langle v_i, v_j \rangle | \langle v_i, v_j \rangle \in G^{gc} \text{ and } x_i \neq x_j\} \quad (8)$$

where $x_i$ and $x_j$ are the labels for nodes $v_i$ and $v_j$, respectively; and the default labels for nodes $s$ and $t$ are $l_1$ and $l_2$, respectively. The cost of the cut $C^{gc}$ is given by

$$E(C^{gc}) = \sum_{\langle v_i, v_j \rangle \in C^{gc}} w(v_i, v_j)$$
$$= \underbrace{\sum_{v_i \in \mathcal{V}} x_i \cdot w(v_i, t) + (1 - x_i) w(v_i, s)}_{data\ term} + \beta \underbrace{\sum_{v_i \in \mathcal{V}, v_j \in N_i} (x_i - x_j)^2 w(v_i, v_j)}_{smoothness\ term} \quad (9)$$

In Eq. (9), $w$ defines the weight of edge in $\mathcal{E}^{gc}$; and the cost function of a cut can be decomposed into two terms: the data term and the smoothness term. The data term is similar to the likelihood energy of MAP-MRF, which is usually modeled based on domain related knowledge (e.g., color distribution, shape and location); and the smoothness term is usually to penalize the discontinuity between neighboring nodes. The segmentation of an image is to find a cut that minimizes $E(C^{gc})$; the minimum *s-t* cut is equivalent to maximize the flow (max-flow) from *s* to *t* according to the Ford-Fulkerson theorem [19]. If a cost function is submodular [20], it can be represented by a graph ($G^{gc}$), and the function can be minimized by using the max-flow algorithm. The Boykov-Kolmogorov version [21] of the implementation of the max-flow algorithm can be downloaded from http://vision.csd.uwo.ca/code/. However, it is not for BUS segmentation; and users need to modify the code accordingly for BUS segmentation.

Applying graph cuts for BUS image segmentation, the key issue is how to define the data and smoothness terms. [22] proposed a fully automatic BUS image segmentation framework in which the cost function modeled the information in the frequency and space domains. The data term modeled the tumor pose, position and intensity distribution; and the weights are given by

$$w(v_i, t) = -log[G(i) \cdot \Pr(d_i | x_i = 1)] \quad (10)$$
$$w(v_i, s) = -log[(1 - G(i)) \cdot \Pr(d_i | x_i = 0)] \quad (11)$$

where $G(i)$ is a 2D elliptical Gaussian function to model the tumor pose and position, and is constructed based on the results of the ROI generation [22]; and $Pr(d_i|x_i)$ defines the intensity distributions of the tumor and nor-tumor regions. The smoothness term is constructed based on the intensity discontinuities in the space domain and the edge in the frequency domain; and the weight function is defined by

$$w(v_i, v_j) = \left(1 - ED(v_i, v_j)\right) + e^{-(d_i-d_j)^2/2\delta^2}/D_e(v_i, v_j) \qquad (12)$$

where *ED* is the edge detector defined in the frequency domain, and $D_e$ is the Euclidean distance between two nodes. For details, refer [22, 23].

In [24], user needs to specify a group of foreground (tumor) regions (*F*) and a group of background regions (*B*) to initialize the graph. The weight of any *t-link* is set to ∞ if the node belongs to *F ∩ B*, and all other weights of *t-links* are set to 0; the weight function of the smoothness term is defined by utilizing region intensity difference and edge strength [138]. In [25], the weights of *t-links* were determined online by training a Probabilistic Boosting Tree (PBT) [26] classifier; and $w(v_i, v_j)$ was learned offline utilizing the training set by the PBT. Hao et al. [27] constructed a hierarchical multiscale superpixel classification framework to define the weights in the data term. In [28], both the region-level [29] and pixel-level features were utilized to define the weights in the data term; and all weights were learned by using a structural SVM [30].

### 3.3 Summary

Graph-based approaches account for the second largest portion of BUS image segmentation literature (Fig. 1). They are among the earliest techniques for BUS image segmentation, fade away due to the successes of other powerful approaches such as deformable models (section 3), and surge again because of the dramatic advances of graph models and energy optimization algorithms.

MRF-MAP-ICM is a flexible framework for image multi-partition (not just tumor or non-tumor for BUS images). Most BUS image segmentation approaches based on this framework achieve good performance by designing better likelihood energy and obtaining better initialization. Only obtaining locally optimal solution is the main shortcoming of these approaches. Graph cuts provides an efficient framework for image bi-partition and MRF energy global optimization, and is the main factor to make graph-based models popular again in BUS image segmentation. The approaches based on graph cuts focus on designing more comprehensive data and smoothness terms to deal with low contrast, inhomogeneity, and not well-defined boundary of BUS images. The "shrink" problem is the common disadvantage of these approaches, and the extension [18] could only find the approximate solution of multiple labeling of graph cuts. Normalized cut (N-Cut) [31] could avoid the problem by considering both the disconnected links and the links inside each component. [27, 32 - 34] applied normalized cut for BUS image segmentation. However, without the data term, N-Cut cannot integrate semantic information into the energy, and usually needs user interaction or combining with other approach to achieve good performance; and its high computational cost is another drawback of N-Cut.

A detailed comparison of seven typical graph-based BUS image segmentation approaches is listed in Table 2. As illustrated in Table 2, two approaches [25, 28] learned all parameters of graph models by using discriminative learning. This strategy enhances the robustness of graph models. Another potentially useful strategy is to build high-order graph models, in which every pair of nodes in the graph has an edge; and it can be applied to represent the global correlation among nodes, and can obtain a more representative smoothness term. An efficient approach to optimize the fully connected graph model was proposed in [36].

Table 2. Comparison of graph-based approaches

| Ref. | Year | Category | F/S | Images | Useful Strategies | Disadvantages | Issues |
|---|---|---|---|---|---|---|---|
| [9] | 1998 | MRF-MAP-ICM | S | 30 | Integrate texture distribution; estimate distribution parameters locally | Depend heavily on parameters estimation; assume the Gaussian distribution of features | 2.3 |
| [13] | 2016 | MRF-MAP-ICM | S | 33 | Integrate tissue stiffness information from ultrasound elastography | Assume the Gaussian distribution of features | 2.3 |
| [23] | 2012 | Graph cuts | F | 184 | No distribution assumption; construct smoothness term in both spatial and frequency domains | The "shrink" problem | 2.3 |
| [24] | 2010 | Graph cuts | S | 13 | Graph cuts on regions | Manually set the foreground and background priors; "shrink" problem | 2.3 |
| [25] | 2010 | Graph cuts | F | 347 | Learn the data and smoothness terms using deterministic model | The "shrink" problem | 2.3 & 2.4 |
| [28] | 2013 | Graph cuts | F | 469 | Integrate features of superpixel and local detection windows; learn all model parameters by using structured support vector machine [30] | The "shrink" problem | 2.3 & 2.4 |
| [32] | 2012 | N-cut | S | 100 | No "shrink" problem | Manually select ROI | 2.3 |

F denotes fully automatic approach and S denotes semi-automatic approach; the column of Issues shows the issues of concern of each approach discussed in Sections 2.1 – 2.4.

## 4. Deformable models

Deformable models (DMs) are curves/surfaces that can move toward to the object boundary under the influence of forces defined on the curve or surface by using the information of image. DMs can deal with biological structures with significant variability, and permit user to integrate expertise to guide the segmentation when necessary; therefore, they have been applied to BUS image segmentation extensively. DMs are proposed by Terzopoulos [37], and then become a popular and active field after the *snake* approach for planar image was proposed by Kass et al. [38]. Generally, DMs can be classified into two categories according to the curve or surface representation during deformation: (1) the parametric DMs (PDMs) and (2) the geometric DMs (GDMs). In PDMs, a curve or surface is represented by its parametric form explicitly. GDMs represent curves and surfaces implicitly as a level set of a scalar function. It can adapt to topological changes of targets, which is helpful for segmenting multi-objects and objects with unknown topology.

**4.1 Parametric deformable models**

Finding object boundaries using PDMs is formulated to minimize an energy function including internal and external energies. The internal term controls the continuity and smoothness of curves/surfaces; and the external energy function is calculated using image features to attract curves to object boundary. Let $C(p) = (x(p), y(p)), p \in [0, 1]$ be a deformable curve; it moves to the optimal object boundary by minimizing

$$\mathcal{E}(C) = \underbrace{\frac{1}{2}\int_0^1 (\alpha \cdot (C')^2 + \beta \cdot (C'')^2) dp}_{\mathcal{E}_{internal}(C)} + \underbrace{\int_0^1 P(C(p)) dp}_{\mathcal{E}_{external}(C)} \tag{13}$$

where $C'$ indicates the first order derivative of $C$ that keeps the continuity of curve, and $C''$ is the second order derivative that controls the smoothness of curve; $\alpha$ and $\beta$ are the weights; large $\alpha$ will lead the change in distances between points on curve to have high cost, and large $\beta$ will penalize more on non-smooth curve. $P$ is the cost function based on image features and the general formulation of $P$ in the original DMs could be found in [38].

The problem of finding the curve $C$ minimizing $\mathcal{E}(C)$ is to find the extrema of functional which satisfies Euler-Lagrange equation [38]:

$$\frac{\delta \mathcal{E}(C)}{\delta C} = -\alpha C'' + \beta C'''' + \nabla P = 0 \tag{14}$$

The above equation states that the functional derivative vanishes at the optimal curve. Given an initial curve $C^0$, we can apply the gradient descent minimization to optimize $\mathcal{E}(C)$ iteratively. At step $t + 1$, the $p$th point on curve $C$ is updated as

$$C^{t+1}(p) = C^t(p) + \gamma F(C^t(p)) \tag{15}$$

where $\gamma$ is the step size, $F$ is the force on curve defined as the negative of the functional derivative

$$F = \underbrace{\alpha(C'') - \beta(C'''')}_{F_{int}} \underbrace{-\nabla P}_{F_{ext}} \tag{16}$$

where the internal forces $F_{int}$ discourage curve stretching and bending, and the external forces $F_{ext}$ often are composed of multiple forces to make the models flexible enough to handle different tasks. The default PDMs discussed above may converge poorly for real image segmentation tasks because of several limitations, e.g., failing to converge to concave boundaries, and poor performance if the initial curve is not close to the minimum. Several variants [39, 70, 146] have been proposed to address the problems of the default PDMs by introducing different external energy.

## 4.2 Geometric deformable models (GDMs)

GDMs [40, 41] are proposed to overcome the two primary limitations of PDMs: lack of topology adaptivity and parameterization dependence. In GDMs, a 2D curve ($C$) at time $t$ is represented implicitly as the *zero-level set* of a scalar function ($\phi$): $C^t = \{C(p,t) = (x,y) | \phi(x,y,t) = 0\}$, and the curve deformation utilizes curve's geometric measures such as the unit norm and curvature, and image properties.

Let $\vec{v}(p,t)$ and $\vec{n}(p,t)$ be the speed and the unit normal vector of the $p$th point of an evolving curve at time $t$, respectively; and the partial derivative of $C$ with respect to $t$ can be defined as the normal component of $\vec{v}$:

$$\frac{\partial C}{\partial t} = \vec{v} \cdot \vec{n}, \quad (17)$$

since the tangential component does not affect the geometry of an evolving curve.

Because of the curve representation by using level set function in GDMs, the deformation of curve is realized by evolving the level set function rather than tracking the evolution of the curve, which enables the automatic topology adaption of the embedded curve. Given a level set function $\phi(x,y,t)$ and a curve $\{C(p,t)\}$, we have

$$\phi(C(p,t),t) = 0, and \ \frac{\partial \phi}{\partial t} + \nabla \phi \frac{\partial \phi}{\partial t} = 0 \quad (18)$$

If $\phi$ is negative inside and positive outside, the inside-pointing unit norm can be defined as $\vec{n} = -\frac{\nabla \phi}{|\nabla \phi|}$, and we can obtain

$$\frac{\partial \phi}{\partial t} = -\nabla \phi \cdot \vec{v} \cdot \vec{n} = \vec{v} \cdot |\nabla \phi| \quad (19)$$

Notice that the speed function $\vec{v}$ is the key of GDMs; and it is usually defined as a function of the curvature ($\kappa$); to avoid the final curve shrinking to a point, image related information such as gradient is usually formulated to slow down the evolving process and to attract the evolving curve to the desired boundary. Three important issues need to consider in implementing GDMs for BUS image segmentation:

(1) Designing the speed function. The foremost step in applying GDM is to design an appropriate speed function that can stop the evolving curve at the desirable object contour. A commonly used speed function is formulated as [42-44]

$$\vec{v} = g \cdot (V_0 + \kappa) - \langle \nabla g, \vec{n} \rangle \quad (20)$$

$$g = \frac{1}{1+|\nabla(G_\sigma * I)|} \quad (21)$$

where $V_0$ is a constant to speed up the curve deformation, $g$ is the stopping function by using image gradient to slow down and to stop the curve evolving at the high gradient locations, and the second term ($\langle \nabla g, \vec{n} \rangle$) makes the stopping power stronger when there are no perfect edges. (2) Initializing the level set function. The initial level set function is usually the signed distance from each point to the zero level set. A fast computation approach is described in [41]. (3) Reinitializing the level set function. The speed function is defined only on the zero level set; the level function deformation requires it to be extended to all level sets; and the extensions [45] can cause the irregularity problem of level set function. The re-initialization is applied to provide a numerical approach to replace the level set function with newly signed distance functions. For more information about the re-initialization, refer [45-47].

**4.3 Edge-based vs. Region-based DMs**

DMs provide a flexible image segmentation framework that can incorporate both low-level image features and various prior knowledge, such as edge, local region statistics, shape, and intensity distribution. The DMs can be classified into edge-based [38, 41, 42, 44, 48] and region-based [47, 49, 66] according to the information to construct the external force in PDMs or speed function in GDMs. Edge-based DMs aim to attract the evolving curve to object boundary by defining the deformation force or speed using image gradient [42-44], and the models depend on image gradient to stop curve deformation. Region-based DMs model region features to guide the curve deformation and can achieve better performance than edge-based models; specially, if images have plenty of noise and weak object boundary.

**4.4 Deformable models for BUS image segmentation**

The application of DMs for BUS image segmentation can be divided into two stages. In the first stage, BUS image segmentation approaches apply the ideas of PDMs, and focus on developing method to generate good initialization. To achieve global deformation and local variation of irregular tumor, Chen et al. [50] applied the B-snake model [51] for BUS image sequence segmentation. In [52], Chen et al. proposed a cell-based dual snake [53] to handle the two problems of applying traditional DMs for BUS image segmentation: (1) initial contour should be placed close to the tumor boundary; and (2) difficulty to capture highly winding tumor boundary. In [54, 55], Madabhushi et al. proposed a fully automatic approach for BUS image segmentation using PDM which is initialized by utilizing the boundary points produced in tumor localization step. In [56, 57], Sahiner applied PDM for 3D BUS tumor segmentation, the external forces had two terms: the first term was defined on image gradient by using 3×3 Sobel filters, and the second term is the balloon force. In [58, 59], Chang et al. applied DM for 3D breast ultrasound image segmentation. The sticks filter [60] was utilized to enhance edge and reduce speckle noise. Huang et al. [61] proposed an automatic BUS image segmentation approach by using the gradient vector flow (GVF) model [70]. The initial boundary was obtained by using Watershed approach. Yap et al. [62] proposed a fully automatic approach by applying GVF model. In its initial contour generation step, a fixed threshold was employed for candidate generation.

In the second stage, many works modified DMs to improve segmentation performance. Yezzi et al. [42] modified traditional GDMs by having an additional term ($\langle \nabla g, \vec{n} \rangle$) that provided stronger stopping power at object edges. Deng et al. [63] proposed a fast GDM method for biomedical image segmentation. It only updated the speed function and evolved the level set functions in a small window. The experimental results showed that it was much faster than the narrow band algorithm [52]. Liu et al. [64] proposed a fully automatic BUS image segmentation based on texture classification and GDM. Gomez et al. [65] proposed a BUS image segmentation approach based on the active contour without edges (ACWE) model [66] which defined the stopping term using Mumford-Shah technique and was robust to segment images with weak boundaries.

Liu et al. [67] proposed an interactive BUS image segmentation approach utilizing region-based GDMs, in which the probability density difference between the intensity distributions (tumor and background) and the estimated Rayleigh distribution was applied to enforce priors of intensity distribution. The approach was compared with two other GDM approaches [66, 68] using 79 BUS images. Rodtook et al. [69] modified the generalized GVF [70] approach based on a continuous force field analysis, and applied it to BUS image segmentation. Daoud et al. [71] considered the SNR and local intensity value as two important features for estimating tumor boundary, and built a two-fold termination criterion based on the two features in discrete dynamic DMs [72]. Gao et al. [73] proposed a level set approach for BUS image segmentation based on the method in [47] by redefining the edge-based stop function using phase congruency [74] which is invariant to intensity magnitude, and integrated GVF model into the level set framework. Cai et al. [75] proposed a phase-based DM in which the local region statistics [49] was introduced to solve the inhomogeneous intensity problem, and the phase-based edge indicator is used to replace the gradient-based edge operator. Lin et al. [76] modified the local region-based level set approach [77] by using additional constrained energies centered at four markers specified by radiologist. [78] proposed a fully automatic robust region-based level set approach with contour points classification (low contrast class and high contrast class). For the points in the low contrast class, both the global and local region-based energies [66, 77] were used; while for the high contrast class, only the local region-based [77] energy was utilized. [79] adopted the region-based approach in [78], proposed a learning based approach (multivariate linear regression and support vector regression) to produce the parameters adaptively. Yuan et al. [80] proposed a new level set based DM approach; the local divergence and a smoothing kernel were introduced to improve the speed function. Kuo et al. [81, 82] utilized user interaction to generate the volume of interest (VOI), applied the radial-gradient index [83] to estimate the initial lesion boundary, and implemented the region-based DM [48, 49] iteratively to find the final contour. The stopping criterion was defined as $(\bar{I}_F^{t+1} - \bar{I}_F^t) - (\bar{I}_B^{t+1} - \bar{I}_B^t) = 0$ [84], where $\bar{I}_F^t$ and $\bar{I}_B^t$ were the mean intensities inside and outside the segmented regions at step t, respectively.

Table 3. Comparison of BUS image segmentation approaches based on deformable models

| Ref. | Year | Category | F/S | Images | Useful Strategies | Disadvantages | Issues |
|---|---|---|---|---|---|---|---|
| [55] | 2003 | PDMs | F | 90 | Use the balloon force to increase the attraction range | Fixed RP; difficult to set the strength of the balloon force | 2.2 & 2.3 |
| [58] | 2003 | PDMs | F | 8(3D) | Define the external force using local texture features | Validated only on a small dataset, and sensitive to initialization | 2.2 & 2.3 |
| [57] | 2004 | PDMS | S | 102 (3D) | Use the balloon force to increase the attraction range | No quantitative evaluation of the segmentation | 2.3 |
| [62] | 2007 | PDMS | F | 360 | Use GVF to extend the attraction range and to handle concave boundaries | Fixed threshold and RP; no quantitative evaluation | 2.2 & 2.3 |
| [64] | 2009 | PDMs | F | 103 | Use a well-trained texture classifier to detect tumor ROI | Predefined rules to exclude false classified regions | 2.2 & 2.3 |
| [67] | 2010 | GDMs | S | 79 | Model the difference between the regional intensity distribution and the estimated prior distribution | Slow; sensitive to initialization | 2.3 |
| [73] | 2012 | PDMs | S | 20 | Redefine the edge-based stopping function using phase information | Validated only on a small dataset; sensitive to noise | 2.3 |
| [75] | 2013 | GDMs | S | 168 | Use both local statistics and phase information to define the speed function; and handle weak boundary and inhomogeneity problems better | Slow | 2.3 |
| [78] | 2013 | GDMs | F | 861 | Region-based GDM and solve the inhomogeneity problem better | Slow and sensitive to initialization | 2.2 & 2.3 |
| [81] | 2014 | GDMs | S | 98(3D) | Use Region-based GDM and handle inhomogeneity problem | Slow and need user interaction to extract VOI | 2.3 |

F denotes fully automatic approach and S denotes semi-automatic approach; the column of Issues shows the issues of concern of each approach discussed in Sections 2.1 – 2.4.

### 4.5 Summary

DM is the most popular approach applied to BUS image segmentation. For PDMs, the explicit curve or surface representation allows direct model interaction and can lead to fast implementation. The results of most PDMs are sensitive to initialization, and different initial curve or surface may converge to different local minimal locations and lead to quite different results; consequently, many variants of PDMs extend the attraction range to avoid local minima. The curves/surfaces of PDMs do not split and merge during the deforming process, which makes PDMs unable to adapt to topological change and to segment multiple objects with single initialization. GDMs apply level set functions to represent curves and surfaces implicitly, and inspire great progress in the related fields. There are two advantages of GDMs: (1) they can allow shapes to change topology during the evolving process, which makes them suitable for segmenting multiple objects and time-varying objects; and (2) the numerical computation of the curve and surface propagation can be implemented without parameterizing the objects. GDMs transfer the *n*-dimensional curve/surface deformation into *n*+1-dimensional level set function, which needs to extend the speed function to all level set functions and increases the computational cost greatly. Table 3 presents a detailed comparison of 10 typical DMs-based BUS image segmentation approaches. Due to the advantages, GDMs become more popular than traditional PDMs in BUS image segmentation; and most successful approaches focus on improving the performance of GDMs to deal with the weak boundary and inhomogeneity of BUS images. There are two useful strategies: (1) redefining the stopping function ($g$), and making it independent of image gradient; and (2) redesigning the speed function by using regional statistics.

## 5. Learning-based approaches

Image segmentation can also be viewed as a classification problem, i.e., classifying pixels or superpixels into different categories. Therefore, it is quite common to apply machine learning approaches to image segmentation tasks. Both supervised and unsupervised learning approaches have been employed in BUS image segmentation.

### 5.1 Unsupervised learning approaches

K-means and Fuzzy C-means (FCM) are two popular unsupervised learning (clustering) approaches. Because K-means can be viewed as a special case of FCM, we only present the theoretical background of FCM in this section. FCM was proposed in [85] and improved in [86]. Let $D = \{d_1, \cdots d_n\}$ be a finite set of data (pixels or superpixels), $LCs = \{ct_i\}_{i=1}^{C}$ be a list of $C$ cluster centers; and FCM partitions set $D$ into $C$ clusters by minimizing the following objective function:

$$\min_{C} \sum_{i=1}^{n} \sum_{j=1}^{C} (u_{ij})^m \|d_i - ct_j\|^2 \tag{22}$$

In Eq. (22), $u_{ij} \in [0,1]$ is the membership value representing the degree of data point $x_i$ belonging to cluster $j$, and is given by

$$u_{ij} = \frac{1}{\sum_{c=1}^{C} \left(\frac{\|d_i - ct_j\|}{\|d_i - ct_c\|}\right)^{\frac{2}{m-1}}} \tag{23}$$

where $m$ ($m \in R$ and $m \geq 1$) is the fuzzifier, and determines the degree of cluster fuzziness; a large $m$ leads fuzzier cluster (smaller $u_{ij}$); if $m = 1$, $u_{ij}$ takes values 0 or 1, which implies a hard partition (K-means); and $m$ is usually set to 2 if no domain knowledge is introduced. The cluster centers are computed on all data points and weighted by their membership values:

$$ct_j = \frac{\sum_{i=1}^{n}(u_{ij})^m d_i}{\sum_{i=1}^{n}(u_{ij})^m} \qquad (24)$$

The objective function (Eq. (22)) is optimized iteratively to find the local minimum in two simple steps: (1) decide the number of clusters ($C$) and assign the initial membership values ($u_{ij}$); and (2) iteratively update the cluster centers (Eq. (24)) and the membership values (Eq. (23)) until the membership values' change between two iterations is less than a predefined threshold.

The main advantage of the FCM is that each data point can belong to every cluster with a corresponding membership value rather than just belongs to one cluster as in K-means, and FCM can achieve better performance for overlapped data points. However, like K-means, the FCM algorithm can only find the local minima and the results depend on the initialization. In [7, 9, 10], K-means was utilized to estimate the parameters of distributions in graph-based models, and the predefined number of clusters should be set. Xu et al. [87] proposed a BUS image segmentation method using the spatial FCM (sFCM) [88] with local texture and intensity features. Lo et al. [89] applied FCM to generate image regions in four clusters, then extracted the morphology, location, and size features of each region; and finally trained a linear regression model [90] to produce the tumor likelihoods for all regions. The region with the highest likelihood is considered as a tumor. Moon et al. [91] applied FCM to image regions produced by using the mean shift method [92]; and [91] trained a linear regression model to estimate the tumor likelihoods of candidate regions utilizing seven quantitative features which were extracted according to the American College of Radiology (ACR) Breast Imaging Reporting and Data System (BI-RADS) [93]. [89, 91] did not discuss how to initialize the membership values for FCM. To deal with the blur boundary and uncertainty in BUS images, Shan et al. [94, 95] extended the FCM and proposed the neutrosophic l-means (NLM) clustering which takes the indeterminacy of membership into consideration, and can handle the uncertainty in BUS images much better.

### 5.2 Supervised learning approaches

*5.2.1 Support vector machine (SVM)*

SVM is one of the most popular supervised-learning models in machine learning, and can be utilized for both linear classification (linear SVM) and non-linear classification (kernel SVM) [96, 99] by mapping its inputs to high dimensional spaces. Let $\{(d_i, x_i)\}_{i=1}^n$ be a training dataset of $n$ points where $x_i$ is either 1 or -1, indicating the class of data point $d_i$; and SVM aims to find a hyperplane which can separate the training samples by a gap as wide as possible. Let $w$ be the normal vector to the hyperplane, and $\{\zeta_i = max(0, 1 - x_i(w \cdot d_i + b))\}_{i=1}^n$ be slack variables for the soft margins; then the problem can be formulated as a constrained quadratic optimization problem [97]

$$\begin{aligned} min \ & \frac{1}{n} \sum_{i=1}^n \zeta_i + \lambda \|w\|^2 \\ s.t. \forall \ & i, x_i(d_i \cdot w + b) \geq 1 - \zeta_i \ and \ \zeta_i \geq 0 \end{aligned} \quad (25)$$

Finally, $w$ and $b$ learned from the training dataset can be used to classify new data by computing $x_i = sgn(d_i \cdot w + b)$. Liu et al. [98] trained a kernel SVM classifier [99] using the local image features to classify small image lattices (16 × 16) into tumor or non-tumor classes; the radius basis function (RBF) was utilized; and 18 features, including 16 features from co-occurrence matrix and the mean and variance of the intensities, were extracted from a lattice. Jiang et al. [100] proposed two-step BUS segmentation approach. First, a set of candidate tumor regions were produced by using Adaboost classifier [101] and 24 Haar-like features [102]; and a SVM classifier was trained using the quantized intensity features produced by K-means clustering to determine the false positive and true positive regions. Second, random walk [173] was applied to generate the final tumor boundary by placing seed at the center of each true region.

*5.2.2 Artificial Neural network (ANN)*

ANN is another popular supervised learning approach for BUS image segmentation. A typical ANN has three layers: an input layer, a hidden layer and an output layer, interconnected by weighted links, e.g., $w_j^I = [w_{1,j}^I, \cdots, w_{nI,j}^I]^T$ is the weight vector of the links between the *j*th hidden units and the input units, and $w_k^O = [w_{1,k}^O, \cdots, w_{nI,k}^O]^T$ is the weight vector between the *k*th unit in the output layer and the hidden units. The units in the layers are usually called 'neurons'. The input neurons represent the feature vector $x = [x_1, \cdots, x_{nI}]^T$; $h_j$ is the output of the *j*th hidden neuron; and $z = [z_1, \cdots, z_{no}]^T$ yielded in the output layers will be used for classification. The output of the *k*th neuron in the output layer ($z_k$) is

$$z_k = \varphi \left( \sum_{j=1}^{nh} w_{k,j}^O \cdot \varphi \left( \left(w_j^I\right)^T \cdot x \right) \right) \quad (26)$$

where $\varphi(\cdot)$ is the activation function of a neuron (in hidden or output layers), and converts a neuron's weighted input to its output value; and the preferred activation function should be non-linear and differentiable such as the sigmoid and hyperbolic tangent functions.

Huang et al. [103] proposed an ANN-based method to segment 3D BUS images by processing 2D image slices. First, thresholding was applied to generate candidate regions; then five region features (average gray level intensity, entropy, ratio of region size to slice size, ratio of region size to the size of its bounding box, and the distance from the region center to image center), were used as the inputs of NN. The number of hidden unites and output units was not discussed. [104] trained an ANN to generate the threshold for BUS image segmentation. Two feature extraction approaches were proposed: 1) using $128 \times n_{key}$ SIFT features where $n_{key}$ is the number of key points; and 2) exacting 4 texture features (contrast, correlation, energy and homogeneity) of a 40×40 region. The ANN has 3 layers, 60 nodes in hidden layer, and one node in the output layer. The stop criterion is $10^{-7}$ of the MSE. Shan et al. [105, 106] trained an ANN using three new features: the phase in the max-energy orientation (PMO) based on phase congruency, radial distance (RD) and the joint probability of intensity and texture [30]. The NN had 6 hidden units and 1 output unit.

### 5.2.3 Naive Bayesian classifiers (NBCs)

NBCs are a family of probabilistic classifiers based on the strong independence assumption: given the class variable, each feature is independent of other features in the feature vector. By using the strong independence assumption and the Bayes' theorem, the conditional distribution over the class variable is

$$p(x_k|d) = \frac{1}{Z} p(x_k) \prod_{i=1}^{n} p(d_i|x_k), k = 1, \cdots, K \tag{27}$$

where $d = [d_1, \cdots, d_n]^T$ is the vector of $n$ independent features, and $x = [x_1, \cdots x_K]^T$ is $K$ class labels; $Z = p(x)$ is a constant if the values of the feature variables are known; and $p(x_k)$ is the prior distribution of $x_k$. NBC commonly combines the conditional distribution and the MAP decision rule to construct the classifier:

$$\hat{x} = \underset{x_k}{\operatorname{argmax}} p(x_k) \prod_{i=1}^{n} p(d_i|x_k) \tag{28}$$

Applying NBC, the first thing is to calculate the prior by assuming equiprobable classes ($p(l_k) = p(l_j), i \neq j$) or by estimating from the training set; and then one must assume the conditional distribution ($p(x_i|l_k)$) over feature variables or learn a nonparametric model from training set.

Yang et al. [107] proposed a whole breast tumor detection method by classifying slice pixels into tumor ($x_1$) or normal tissue ($x_2$) by using NBC. Two features, local ($5 \times 5$ mask) intensity mean and stick filter [60] output, were utilized; the class priors were assumed to be equiprobable, and the conditional distribution of each feature ($x_i$) was assumed to be Rayleigh distribution:

$$p(x_i|l_k) = \frac{x_i}{\sigma_R} e^{-x_i^2/2\sigma_k^2} \quad , k = 1, 2 \tag{29}$$

where $\sigma_k$ is the Rayleigh parameter and can be estimated from training data. NBC produced a set of suspected lesions, and a two-phase lesion selection method based on region shape features and region continuity and volume size were applied for final tumor region decision [108].

Table 4. Comparison of learning-based BUS image segmentation approaches

| Ref. | Year | Category | F/S | Images | Advantages | Disadvantages | Issues |
|---|---|---|---|---|---|---|---|
| [7, 9, 10] | 95, 98, 2003 | Adaptive K-means | S | < 10 | Achieved better performance than the standard K-means on images with local intensity variations | Sensitive to initialization; only used for estimating distribution parameters | 2.3 |
| [89] | 2014 | FCM | F | 58 | High sensitive rate | Sensitive to initialization; high false positive rate | 2.2 & 2.3 |
| [91] | 2014 | FCM | F | 148 | High sensitive rate | Sensitive to initialization; used fixed threshold | 2.2 & 2.3 |
| [98] | 2010 | SVM | F | 112 | Utilized local texture features to classify local lattices and achieved high precision and recall ratio | Only produced rough tumor boundaries; depended on post processing rules to refine the results | 2.2 & 2.3 |
| [100] | 2012 | SVM | F | 112 | Balanced sensitivity and specificity | Slow; depended on random walk to generate the final boundary | 2.2 & 2.3 |
| [103] | 2008 | ANN | F | 93 (3D) | Fully automatic | Depended on fixed threshold to produce candidate regions; relatively low performance | 2.2 & 2.3 |
| [106] | 2012 | ANN | F | 120 | Achieved good performance by using the phase information, radial distance and the joint distribution of texture and intensity | Depended on fixed reference point to generate the initial ROI | 2.2 & 2.3 |
| [107] | 2012 | NBC | F | 31 | Achieved high sensitive ratio | Depended on the assumption of intensity distribution; depended on post selection to reduce FPR | 2.2 & 2.3 |

F denotes fully automatic approach and S denotes semi-automatic approach; the column of Issues shows the issues of concern of each approach discussed in Sections 2.1 – 2.4.

## 5.3 Summary

Unsupervised learning is simple and fast, and has been widely utilized in many BUS image segmentation approaches. However, because of the challenging nature of BUS image segmentation, unsupervised approaches are only employed as a preprocessing step to generate candidate image regions and more sophisticated methods are usually employed to perform the final segmentation; for example, in [7, 9, 10], K-means was utilized to estimate the initial parameters of intensity or texture distributions. Supervised learning provides a good framework to integrate different levels of features and to learn the knowledge between the inputs and target outputs. Many BUS image segmentation approaches achieve good performance by using supervised learning approaches. Most of them design features using domain knowledge (feature engineering) to improve the performance. They can be integrated with other segmentation techniques, e.g., in [25, 28], supervised learning approaches learned the parameters of the graph-based models from the training data; and they also can be used to perform segmentation alone, e.g., in [120], a well-trained ANN was applied to perform tumor segmentation. One common disadvantage of the supervised learning approaches is that they cannot produce accurate tumor boundary, and refinement is usually necessary. Table 4 presents a detailed comparison of 10 learning-based BUS image segmentation approaches.

Learning-based approaches thrive in BUS image segmentation in the last decade and we believe new deep learning techniques [162] such as deep convolutional neural networks (CNNs) and recurrent neural network (RNN) will make great progress in segmenting BUS images in the near future. For more details about applying deep learning to medical image analysis, refer the survey paper [171].

**6. Classical Approaches: thresholding, region growing, and watershed**

In this section, we will discuss some classical segmentation approaches applied to BUS image segmentation, and they are usually combined with other methods to achieve good performance.

**6.1 Thresholding**

Thresholding is the most intuitive, simple and fast segmentation approach, and enjoys the popularity in BUS image segmentation. It groups image pixels directly into regions by using a single threshold (two classes) or multiple thresholds (multiple classes) based on pixel features (e.g., intensity, color, local mean, standard deviation, etc). Let *th* be the threshold that segments image pixels into two classes. When *th* is a constant over entire image, the method is called global thresholding; if *th* is changing over the local features, the method is referred as adaptive/local thresholding. Global thresholding is fast and works well when the intensity distributions of objects and background are sufficiently distinct; however, if the object-background contrast is low, image is noisy, and/or illumination varies across the image, global thresholding cannot achieve good performance. However, global thresholding is often used as a pre-segmentation step in BUS image segmentation. There are three main approaches to select the global threshold. The first approach is to choose an empirical value as the threshold for the whole dataset [109-111]; the second approach is to select the threshold for each image based on domain related rules [112, 113]; and the third is to generate the threshold automatically based on statistical-decision theory [22, 114, 115].

**6.2 Region growing**

Region growing extracts regions from a set of pixels (*seeds*) and grows seeds to bigger regions utilizing pre-defined growth criteria. Seed generation: the seeds can be placed by user interactively [116, 117] or generated automatically [54, 55, 113, 118]. [54, 55] selected a seed ($p^*$) automatically from a set of candidate pixels by formulating empirical rules [55]:

$$p^* = \underset{p}{\mathrm{argmax}} \left( \frac{\Gamma(i_p, t_p) \cdot \mathcal{I}_p \cdot Y_p}{d_p} \right) \tag{30}$$

where $i_p$ and $t_p$ are the intensity and texture values of pixel $p$, respectively; $\Gamma(i_p, t_p)$ is pixel $p$'s value of joint intensity and texture probability; $\mathcal{I}_p$ refers to the local mean value of $\Gamma$ around $p$; $Y_p$ is the row position (origin at the lower-left of image) of $p$ and avoids selecting seed from the shadowing region appearing at the bottom of BUS image; and $d_p$ is the distance between $p$ and image center. [118] used this method for seed generation. For each pixel $p$ adjacent to the seed region, if $p$ satisfied (growing criterion): $\mathcal{I}_p/\mathcal{I}_{p^*} \in [\beta_1, \beta_2]$; $p$ will be added to the seed region, where $\beta_1$ and $\beta_2$ are selected by experiment. The growing process will stop until no more pixel satisfying the condition

Shan et al. [113] proposed another automatic seed generation approach. Thresholding was used to generate a group of candidate regions; and the region ranking criteria based on region location, size, and local feature, were utilized to determine a true tumor region ($r^*$):

$$r^* = \underset{r}{\mathrm{argmax}} \left( \frac{\sqrt{A(r)}}{d_r \cdot var(r_c)} \right) \tag{31}$$

where $A(r)$ is the number of pixels in region $r$; $d_r$ is the distance between the center of $r$ and a fixed reference point (center of the top half of image); $r_c$ is the center of region $r$; and $var(r_c)$ is the local variance of a circular region at the center of region $r$; and a pixel inside region $r^*$ will be selected as the seed. Let $I(p)$ and $\bar{I}(p)$ be intensity and local mean intensity of pixel $p$, respectively, the growing criterion is defined by

$$1 - e^{-\frac{var(p)}{100}} \leq t_1 \; and \; 1 - \frac{|I(p) - m(r_s)|}{m(r_s)} \geq t_2 \; or$$

$$1 - e^{-\frac{var(p)}{100}} \geq t_1 \; and \; 1 - \frac{|\bar{I}(p) - m(r_s)|}{m(r_s)} \geq t_3 \tag{32}$$

where $m(r)$ is the average intensity of the current seed region, and $t_1$, $t_2$, and $t_3$ are set as 0.5, 0.2 and 0.99, respectively. The growing processing stops when no more pixel satisfies the above condition.

Kwak et al. [117] defined the cost of growing a region by modelling common contour smoothness and region similarity (mean intensity and size):

$$J(r_s, r) = \frac{(m(r_s) - m(r))}{\alpha \cdot LC(r_s, r) + \beta} \cdot \frac{A(r_s) \cdot A(r)}{A(r_s) + A(r)}, r \in N(r_s) \tag{33}$$

where $m(\cdot)$ denotes the mean intensity of a region, $A(\cdot)$ is the pixel number of the region, $LC(r_s, r)$ is the length of the common contour between the seed region $r_s$ and region $r$, $N(r_s)$ is a set of regions adjacent to $r_s$, and $\alpha$ and $\beta$ are two predefined constants. The region with the minimum value of $J$ will be added to the seed region. The growing repeats until $\sum_{r \in N(r_s)} J(r_s, r)$ over the length of contour $r_s$ reaches the maximum.

**6.3 Watershed**

Watershed is a powerful image segmentation method, and usually produces more stable results than thresholding and region growing approaches. There are different definitions of watershed [119, 120]. The most popular definition is the watershed by flooding [119]. The most common implementation of watershed for image segmentation can be found in [121]. The key issue in watershed segmentation is the marker selection. One approach is to choose the local minimum gradient as the marker, which will usually result in over-segmentation due to noise, and further step such as region merging should be involved. The other approaches choose makers based on more complex predefined criteria that can utilize the task-related priors.

Table 5. Comparison of classical approaches

| Ref. | Year | Category | Purpose | Threshold/Seed/Marker Generation | Additional Comments |
|---|---|---|---|---|---|
| [62, 111] | 2007 2008 | Global thresholding | Pre-segmentation | Fixed threshold | Depended on image enhancement; could not adapt to variations of image quality. |
| [113] | 2008 | Global thresholding | Candidate tumor region generation | Iterative thresholding by finding the local minima of histogram. | Depended on post empirical rules to refine the candidate regions; used fixed reference position. |
| [115] | 2014 | Global thresholding | ROI generation | Otsu's algorithm [130] | Depended on image preprocessing. |
| [55] | 2003 | Region grow | Pre-segmentation | Selected seed by formulating empirical rules. | Used fixed reference position (image center). |
| [117] | 2005 | Region grow | Final segmentation | User interaction to set an elliptical seed region | Depended on image enhancement; the growth criteria were defined by formulating contour roughness and region inhomogeneity. |
| [122] | 2004 | Watershed | Final segmentation | Selected marker by using intensity and connectivity | Depended on image enhancement. |
| [124] | 2010 | Watershed | Pre-segmentation | Decided the external and internal markers by computing the Beucher gradient [131] of the morphological dilation and erosion of the binary image | Depended on image preprocessing; select 255 groups of markers by thresholding image using thresholds from 0 to 255; needed additional geometrical measure [132, 133] to decide the final tumor contour. |
| [127] | 2010 | Watershed | Pre-segmentation | Used regions on the binary edge map as markers | Used empirical rules to refine the results |
| [128] | 2014 | Watershed | Pre-segmentation | Local intensity minima | Used empirical rules to refine the results |

Huang et al. [122] applied the watershed to segment the preprocessed BUS images, and the markers were selected based on grey level and connectivity. [109, 123] used watershed to segment ROI into small regions, and used the predefined criteria (area, mean intensity, geodesic center, etc) to determine the final tumor region. 255 groups of markers were selected by thresholding ($th$ = 1, 2, ⋯, 255) the image [124, 125]; the external and the internal markers were defined by using the morphological dilation and erosion. Watershed method was applied to generate 255 potential lesion boundaries by using the markers on different thresholds; the average radial derivative (ARD) function [132, 133] was applied to determine the final tumor boundary. Zhang et al. [126, 127] applied watershed to determine the boundaries of gray level images. The markers were set as the connected dark regions. [128] applied watershed to generate meaningful regions, and refined the regions by removing the top 50% hyper-echogenic (bright) regions and the regions connected to the image border to generate candidate tumor regions; and the candidate regions were distinguished between tumors and non-tumors by using a logistic regression classifier [129] trained using region morphology, intensity and texture features.

**6.4 Summary**

In this section, we present the theoretic background of three classical image segmentation approaches: thresholding, region growing, and Watershed; and discuss their applications to BUS image segmentation. Table 5 gives a detailed comparison of 10 BUS image segmentation methods. The three approaches are quite simple, fast, and efficient to conduct initial segmentation of BUS image, and facilitate further segmentation procedures. To achieve good performance of BUS image segmentation, two additional steps are usually needed: first, image preprocessing step is employed to improve image quality by denoising and enhancing contrast; second, more delicate approaches are utilized to refine the segmentation results.

**7. Other approaches**

Beside the four main categories of BUS image segmentation approaches discussed in sections 2-5, there exist some interesting approaches presented in few papers. We discuss them briefly in this section.

**Cellular automata (CA)**: CA was introduced by von Neumann [134] and applied to interactive image segmentation [135]. In image segmentation, a cell is usually associated with a pixel or superpixel. A CA is defined as a triplet $CA = (St, \mathcal{N}, \delta)$ where $St$ is the state set, $\mathcal{N}$ denotes the neighborhood system, and $\delta$ is the transition function which defines the rule of updating the cell state based on the states of the neighborhood cells in the previous step. Liu et al. [136] constructed the transition function by integrating the global information on the transition chain and local texture correlation. There are three main advantages of CA-based approaches: (1) support multiple objects segmentation; (2) can generate precise object boundary and do not have the "shrink" problem; and (3) support user input online. These approaches start with user interaction to initialize seed cells, and then update the states of all other cells according to the evolution rule until reaching the stable states or the fixed number of iterations. The computation cost is high; especially, when the image size is large. For a fast CA-based segmentation approach, refer [137].

**Cell competition**: Chen et al. [138] proposed a cell-competition approach for BUS image segmentation. The cells are small image regions generated by using a two-pass Watershed segmentation; and then adjacent cells compete to generate new regions by splitting or merging. There are two types of competitions. In Type I competition, two adjacent cells from different regions compete; a cell may split from a region and merge into another region. In Type II competition, one cell splits from a multi-cell region and forms a single-cell region. Cheng et al. [139] applied the approach to an initial slice selected by user, and used the results to partition the cells of other slices into object or background regions. Chiang et al. [140] extended the approach to segment 3D BUS image and applied graph cuts for finding the final tumor boundary. No task-related knowledge is integrated in the competition mechanism. It is simple and fast, but needs large amount of user interactions to select ROI before the competition or to select tumor regions after the competition.

**Radial gradient index (RGI)**: RGI [141] calculates around the boundary of each candidate partition, and the partition with the largest RGI value was selected as the tumor region.

$$RGI(\mathcal{M}_i) = \frac{1}{\sum_{p \in \mathcal{M}_i}|\nabla f(p)|} \sum_{p \in \mathcal{M}_i} \nabla f(p) \cdot \frac{\vec{r}(p)}{|\vec{r}(p)|} \qquad (34)$$

In Eq. (34), $\mathcal{M}_i$ is a set of contour points of the *i*th image partition; $|\nabla f(p)|$ is the absolute value of the intensity gradient at point *p*; and $\vec{r}(p)$ is the radial vector from the partition center to point *p*. The RGI value measures the proportion of the intensity gradients of the boundary points along the radial direction. It takes values in [-1, 1]; RGI value 1 indicates that all the gradients point outward along the radial vector; and -1 signifies that all the gradients point inward along the radial vector. For BUS tumor regions, the RGI values are expected to close to 1. The application of RGI for BUS image segmentation can be found in [142 - 146]. The RGI calculation is simple and easy to implement; however, it calculates a group of RGI values for each pixel, and the computation cost is quite high; furthermore, it depends on image gradient, and cannot obtain accurate tumor boundary of BUS image due to low image quality.

## 8. Conclusions and Future Directions

In this paper, we have reviewed the automatic BUS image segmentation approaches. A general comparison between the categories of BUS image segmentation approaches is listed in Table 6. The major future directions of BUS image segmentation are summarized as follows:

**Unconstrained BUS image segmentation techniques**: currently, most BUS image segmentation approaches work well on BUS images collected in controlled settings such as high image contrast, less artifacts, containing only one tumor per image, etc. However, their performance degrades greatly with BUS images having large variations in image quality, degree and location of artifacts, and number of tumors per image.

Therefore, to promote the application of BUS CAD systems in clinical practice, it is crucial to develop unconstrained BUS segmentation techniques which are invariant to image settings. Potential directions are to learn invariant and discriminative representations of tumors in BUS images, and to model human vision mechanism [110, 154, 176].

**Benchmark**: A publicly accessible BUS image benchmark can be useful for comparing existing approaches, for discovering novel strategies that can contribute to better segmentation performance, for helping researchers to develop better approaches, and for promoting the development and advance of breast cancer research. Building of a publicly accessible BUS image dataset requires incredible effort (many years of hard work, and large amount of resources); however, its impact will be significant and profound.

Table 6. General Comparison of the four main categories

| Categories | | Advantages | Disadvantages |
| --- | --- | --- | --- |
| Graph-Based | MRF-MAP approaches | Organize BUS knowledge and image information in a unified framework; multiple-object segmentation | Only obtain local optimal solution; Sensitive to initial parameter estimation; inefficient optimization algorithm |
| | Graph Cuts | Organize BUS knowledge and image information in a unified framework; efficient optimization algorithm | The "shrink" problem |
| DMs | PDM | Intuitive; fast | Cannot adapt to segment multiple tumors; sensitive to initialization |
| | GDM | Adapt to multiple tumors in an BUS image | Sensitive to initialization; slow |
| Learning-based | Unsupervised learning | Simple; fast | Sensitive to initialization; only applied in the BUS image preprocessing step |
| | Supervised learning | Integrate different levels of features and learn the knowledge between the inputs and target outputs | refinement is usually required |
| Classical approaches | | Simple; fast | Utilized for initial BUS image segmentation; Sensitive to image noise; refinement is required |

**Deep learning**: in the last several years, deep learning has demonstrated impressive performance for many tasks such as object recognition [158], image classification [159], semantic segmentation [160], facial expression recognition [163], speech recognition [164], medical applications [161, 170], etc. Deep learning models have great potential to achieve good performance for BUS image segmentation because of their ability to learn compact image representation using sufficiently large BUS image dataset.

**High performance segmentation**: segmentation performance is usually evaluated by memory cost, speed, and accuracy. Currently, many BUS image segmentation approaches are computation and memory intensive, which limits their widespread applications. For example, it is difficult to integrate resource-intensive algorithms into portable BUS devices for real time applications. In some resource-limited regions or countries, many lives were lost because of unavailability of accurate and low-cost breast cancer detection techniques and devices; high performance approaches consume much less resources than traditional approaches, and is vitally important to provide an affordable means for early detection of breast cancer.

In addition, we have tried to collect codes or software for BUS image segmentation in the public domain; however, we could not find any yet. It may be a future task for the research society.

[121] S. Beucher, F. Meyer, The morphological approach to segmentation: the watershed transformation, Optical Engineering-New York-Marcel Dekker Incorporated, 34 (1992) 433-433.

[122] Y.-L. Huang, D.-R. Chen, Watershed segmentation for breast tumor in 2-D sonography, Ultrasound Med. Biol., 30 (2004) 625-632.

[123] Y. Ikedo, D. Fukuoka, T. Hara, H. Fujita, E. Takada, T. Endo, T. Morita, Computer-aided detection system of breast masses on ultrasound images, in: SPIE MI, 2006, pp. 61445L1-61445L8.

[124] W. Gómez, L. Leija, A. Alvarenga, A. Infantosi, W. Pereira, Computerized lesion segmentation of breast ultrasound based on marker-controlled watershed transformation, Med. Phys., 37 (2010) 82-95.

[125] W. Gómez, L. Leija, W.C.A. Pereira, A.F.C. Infantosi, Segmentation of Breast Nodules on Ultrasonographic Images Based on Marke d-Controlled Watershed Transform, Computación y Sistemas, 14 (2010) 165-174.

[126] M. Zhang, Novel approaches to image segmentation based on neutrosophic logic, PhD thesis, USU, 2010.

[127] M. Zhang, L. Zhang, H.-D. Cheng, Segmentation of ultrasound breast images based on a neutrosophic method, Opt. Eng., 49 (2010) 117001-117012.

[128] C.M. Lo, R.T. Chen, Y.C. Chang, Y.W. Yang, M.J. Hung, C.S. Huang, R.F. Chang, Multi-Dimensional Tumor Detection in Automated Whole Breast Ultrasound Using Topographic Watershed, IEEE TMI, 33 (2014) 1503-1511.

[129] D. W. Hosmer, and S. Lemeshow, Applied logistic regression: Wiley-Interscience, 2000.

[130] N. Otsu, A threshold selection method from gray-level histograms, Automatica, 11 (1975) 23-27.

[131] J.-F. Rivest, P. Soille, S. Beucher, Morphological gradients, in: SPIE/IS&T Symposium on Electronic Imaging: Science and Technology, 1992, pp. 139-150.

[132] K. Horsch, M.L. Giger, L.A. Venta, C.J. Vyborny, Automatic segmentation of breast lesions on ultrasound, Med. Phys, 28 (2001) 1652-1659.

[133] K. Horsch, M.L. Giger, L.A. Venta, C.J. Vyborny, Computerized diagnosis of breast lesions on ultrasound, Med. Phys, 29 (2002) 157-164.

[134] J. V. Neumann, Theory of Self-Reproducing Automata: University of Illinois Press, 1966.

[135] V. Vezhnevets, V. Konouchine, GrowCut: Interactive multi-label ND image segmentation by cellular automata, in: proc. of Graphicon, Citeseer, 2005, pp. 150-156.

[136] Y. Liu, H. Cheng, J. Huang, Y. Zhang, X. Tang, An effective approach of lesion segmentation within the breast ultrasound image based on the cellular automata principle, JDI, 25 (2012) 580-590.
**37 / 40**